\documentclass[fleqn,10pt]{wlscirep}
\usepackage[utf8]{inputenc}
\usepackage[T1]{fontenc}
\usepackage{soul}
\usepackage{multirow}
\usepackage{cleveref}
\usepackage{authblk}
\usepackage[symbol]{footmisc}

\title{YOLO for Knowledge Extraction from Vehicle Images: A Baseline Study}

\author[1,*]{Saraa Al-Saddik}
\author[1,*]{Manna Elizabeth Philip}
\author[1,2,+]{Ali Haidar}
\affil[1]{UNSW, Sydney, Australia }
\affil[2]{NSW Police Force, NSW Police Force Headquarters, Parramatta, Australia}
\affil[*]{\textit{These authors contributed equally to this work}}
\affil[+]{\textit{corresponding author: haid1ali@police.nsw.gov.au}}
\date{\today}

\begin{document}

\flushbottom

\begin{abstract}
Accurate identification of vehicle attributes such as make, colour, and shape is critical for law enforcement and intelligence applications. This study evaluates the effectiveness of three state-of-the-art deep learning approaches YOLO-v11, YOLO-World, and YOLO-Classification on a real-world vehicle image dataset. This dataset was collected under challenging and unconstrained conditions by NSW Police Highway Patrol Vehicles. A multi-view inference (MVI) approach was deployed to enhance the performance of the models' predictions. To conduct the analyses, datasets with 100,000 plus images were created for each of the three metadata prediction tasks, specifically make, shape and colour. The models were tested on a separate dataset with 29,937 images belonging to 1809 number plates. Different sets of experiments have been investigated by varying the models sizes.
A classification accuracy of 93.70\%, 82.86\%, 85.19\%, and 94.86\% was achieved with the best performing make, shape, colour, and colour-binary models respectively. It was concluded that there is a need to use MVI to get usable models within such complex real-world datasets. Our findings indicated that the object detection models YOLO-v11 and YOLO-World outperformed classification-only models in make and shape extraction. Moreover, smaller YOLO variants perform comparably to larger counterparts, offering substantial efficiency benefits for real-time predictions. This work provides a robust baseline for extracting vehicle metadata in real-world scenarios. Such models can be used in filtering and sorting user queries, minimising the time required to search large vehicle images datasets.
\end{abstract}
\maketitle

\section{Introduction}
\begin{figure}[!ht]
\centering
\includegraphics[width=\linewidth]{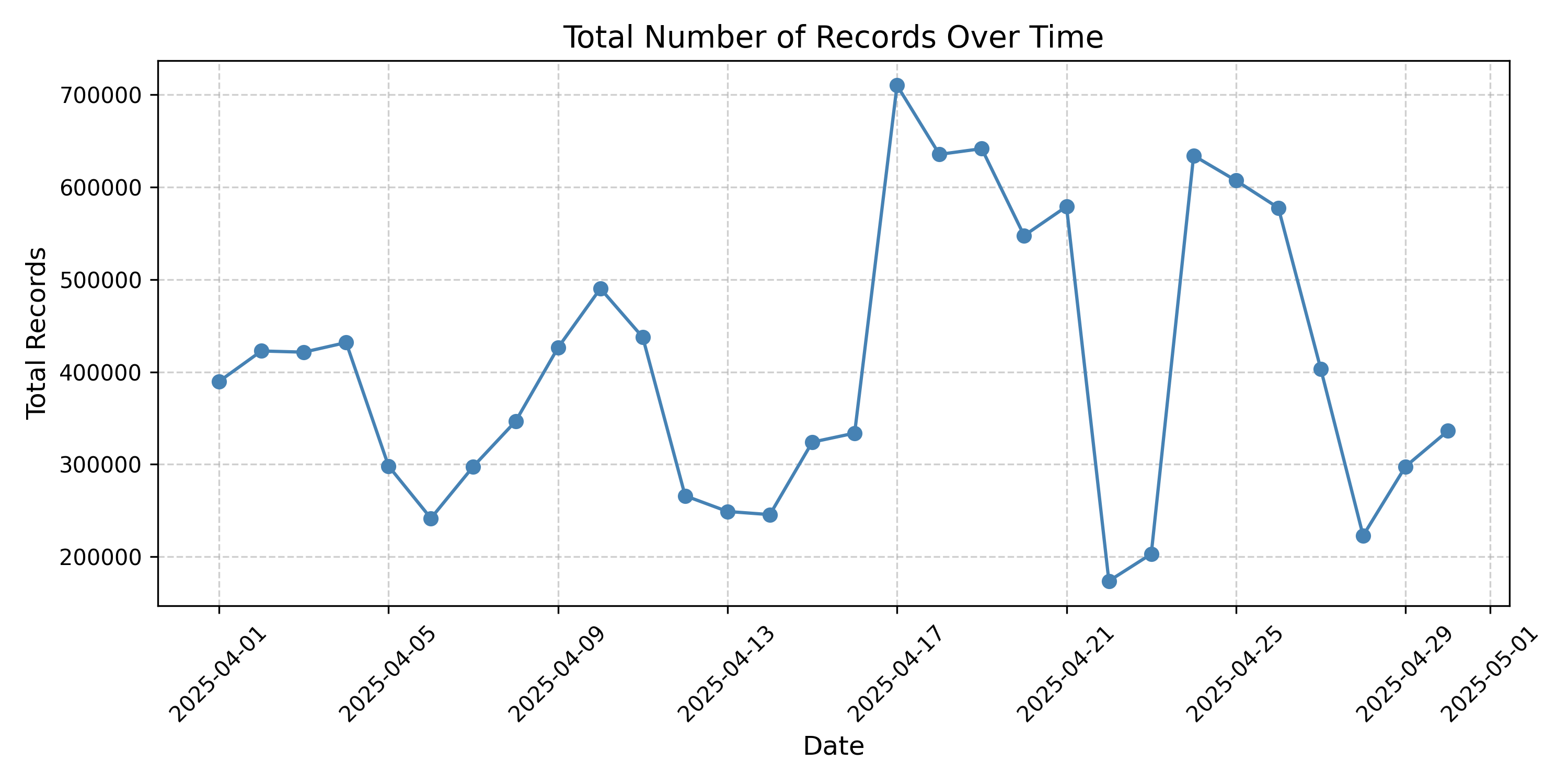}
\caption{Total number of MANPR records per day in April 2025.}
\label{fig:records_over_time}
\end{figure}

The New South Wales Police Force (NSWPF) highway patrol vehicles are equipped with special cameras capable of detecting and reading vehicle number plates. This system, introduced in late 2009, is known as the Mobile Automatic Number Plate Recognition (MANPR) system. The MANPR system works by scanning the surrounding number plates and associated vehicles. For each record, the extracted number plate text, number plate image, vehicle image, time and location details are stored. A large dataset with millions of vehicle images has been established, with hundreds of thousands of new records processed daily. The total number of records processed daily in April 2025 is shown in \autoref{fig:records_over_time}, highlighting the high number of images processed within the system.

\begin{figure}[!ht]
\centering
\includegraphics[width=\linewidth]{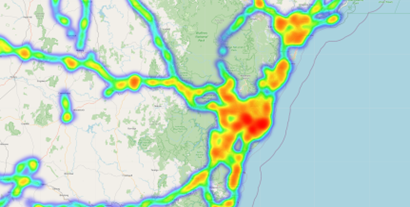}
\caption{MANPR coverage on a single day across NSW.}
\label{fig:manpr_coverage}
\end{figure}

The MANPR system covers the whole state, as shown in \autoref{fig:manpr_coverage}. The MANPR data are used in investigating serious crimes, including murder, kidnapping, public place shootings, and arson during the bushfire season. NSW police force employees with the right access details can use the dataset for different tasks. One common use case involves searching the MANPR dataset through a secure web-based interface, where investigators can retrieve vehicle records by entering a number plate.

In real-world investigations, it is often necessary to perform complex user queries, such as searching for vehicles based on attributes like make, colour, or shape, even when specific number plate details are unknown or unavailable.
However, the current search capabilities are limited and officers cannot search an area for a vehicle based on its metadata such as make, colour, and shape. For example, a search for a 'red' vehicle in a selected area is still not possible. Since not all the vehicles metadata are available, the user must go through all the images in a searched area to find the targeted image if it was captured by the MANPR system. In addition, other metadata queries related to vehicles make and model, shape, damaged parts, accessories, and text on vehicles are still not possible.

The MANPR dataset contains valuable visual information that, if properly mined, could reveal patterns and insights beneficial to policing operations. Although the data exists, officers and analysts are unable to retrieve relevant images based on these descriptive attributes, limiting the effectiveness of data-driven investigations and image-based vehicle tracking.

An approach to addressing this limitation is to use the recognised number plates to retrieve vehicle metadata such as make, shape, or colour from national vehicle registration databases \cite{nazemi_real-time_2019}. However, this is not always feasible with large records required for analysis. Bulk searches to such databases are restricted due to various constraints. 

With the emergence of intelligent transportation systems (ITS), different artificial intelligence (AI) techniques have been proposed and evaluated across vehicle metadata tasks such as vehicle type recognition (VTR), vehicle logo recognition (VLR), vehicle make recognition (VMR), vehicle make and model recognition (VMMR), and vehicle detection \cite{tan_artificial_2024,deshmukh_swin_2023,surwase_multi-scale_2023,gayen_two_2024}. Specifically, in the domain of deep learning, AI systems have shown strong performance in various computer vision applications since the release of AlexNet, the artificial neural network (ANN) that achieved state-of-the-art results in the ImageNet challenge in 2012 \cite{krizhevsky_imagenet_2012}. However, identifying vehicle metadata remains challenging due to the high degree of inter-class similarity. That is, many vehicles from different makes or models share similar designs and visual patterns, making it difficult to distinguish between them reliably \cite{ma_ai-based_2020}. Nevertheless, in cases where number plate information is unclear or unavailable, these AI-based systems offer an effective alternative by enabling the extraction of descriptive attributes- such as vehicle make, shape, and colour directly from the image. 

The problem of identifying vehicle attributes such as make, shape, and colour can be originally framed as an image classification task. In image classification, the objective is to assign a single label to the entire image, assuming that the object of interest dominates the frame. The task of classifying vehicle metadata relies on establishing rich and representative features that allow the distinction between different classes or labels. Most studies in this area have adopted pre-trained ANN architectures as the backbone of their systems. A pre-trained model refers to a neural network trained on a large general-purpose dataset, such as ImageNet, to learn discriminative visual features. These models are then adapted to specific downstream tasks like vehicle metadata classification. Common pre-trained architectures used in this domain include AlexNet, TResNet-L, SE-ResNetXt50, EfficientNet B3, ResNet101, VGG19, Xception, and DenseNet201 \cite{deshmukh_swin_2023,gayen_two_2024, amirkhani_deepcar_2022}. In support of these efforts, several large-scale datasets have been developed for vehicle-related tasks \cite{amirkhani_deepcar_2022,yan_exploiting_2017,tang_cityflow_2019,liu_deep_2016,bai_group-sensitive_2018,liu_deep_2016-1,zapletal_vehicle_2016}. These include the Stanford Cars dataset, which contains 196 vehicle classes, as well as VMMRdb, CompCars, and the MIT Traffic dataset.

Traditional supervised learning methods for vehicle metadata classification involve training a model on a labelled dataset, where each image is annotated with a single class label. Convolutional Neural Networks (CNNs) have been extensively used for this purpose due to their ability to learn rich visual features. Numerous studies have demonstrated high performance on benchmark datasets using a range of CNN architectures.

Early work employed the AlexNet architecture to classify vehicle make and model in the CompCars Surveillance and On-Road Vehicle datasets ~\cite{yang_large-scale_2015,nazemi_unsupervised_2018}. Subsequently, the GoogleNet was shown to outperform the AlexNet on the same benchmark, achieving higher classification accuracy~\cite{szegedy_going_2015}. Similarly, Lee et al. proposed a residual SqueezeNet model that achieved 96.3\% accuracy on a dataset comprising 766 vehicle models ~\cite{lee_real-time_2019}. Liu et al. introduced a multi-layer feature aggregation strategy that utilised outputs from various depths within a CNN backbone to enhance metadata classification ~\cite{liu_learn_2023}.

The Stanford Cars dataset, containing 196 classes of vehicle makes and models, has also been extensively used in this domain ~\cite{krause_3d_2013}. Krause et al. reported a classification accuracy of 97.10\% using a fine-tuned deep CNN model ~\cite{krause_3d_2013}. A broader comparison of CNN variants including VGG, ResNet, and DenseNet architectures on various benchmarks can be found in recent surveys by Tan et al. and Gayen et al.~\cite{tan_artificial_2024,gayen_two_2024}.

To address specific real-world challenges, several methods have been proposed to improve robustness under variable conditions such as night-time imaging. Yu et al. introduced a model for vehicle make and model recognition under night-time conditions, achieving 98.56\% accuracy on a dedicated night-time vehicle dataset ~\cite{yu_night-time_2024}. Other work has explored part-based classification strategies. For example, Amirkhani et al. proposed a framework that trained four neural networks, each focusing on a specific vehicle component (e.g., grille, headlights), and combined predictions using majority voting ~\cite{amirkhani_deepcar_2022}. Their dataset included 20,205 training images and 19,980 testing images from 480 vehicle classes. The framework achieved 96.72\% accuracy using manually annotated bounding boxes, and 92.14\% when bounding boxes were automatically detected.

In addition to using standard CNN backbones, researchers have also proposed integrating custom modules into pre-trained networks to improve fine-grained classification. Ma and Boukerche introduced a Recurrent Attention Unit (RAU) into DenseNet50 and DenseNet101 to enhance the model’s ability to focus on subtle visual differences between vehicle models ~\cite{ma_ai-based_2020}. Their method was evaluated on the Stanford Cars, CompCars, and CompCars Surveillance datasets, achieving accuracies of 93.81\%, 97.84\%, and 98.90\%, respectively. Similarly, Tan et al. proposed a Coarse-to-Fine Context Aggregation (CFCA) module designed to refine contextual representations in the upper CNN layers ~\cite{tan_coarse--fine_2023}. They integrated this module into four backbone architectures VGG16, InceptionV3, ResNet50, and DenseNet169 ~\cite{simonyan_very_2014,szegedy_rethinking_2016,he_deep_2016,huang_densely_2017}. The approach was evaluated across five datasets: CompCarsWeb, Stanford Cars, Car-FG3K, CompCars Surveillance, and Mohsin-VMMR~\cite{wu_graph_2022,ali_vehicle_2022}. CFCA-enhanced ResNet50 yielded the best performance across most datasets, and the authors demonstrated consistent gains over baseline models.

Fine-grained image classification refers to the task of distinguishing between visually similar subcategories within a broader category, such as different vehicle makes and models. These tasks are particularly challenging due to high intra-class similarity and subtle inter-class differences. Approaches in this area often involve identifying discriminative parts of an object, either through part-based CNNs or transformer-based architectures~\cite{he_transfg_2022}.

Nazemi et al. proposed a fine-grained system, ORV-MMR, to classify 50 vehicle categories in the Iranian On-Road Vehicle (IROV) dataset, which includes 90,008 images and 1,117 test samples ~\cite{nazemi_real-time_2019}. They also used the CompuCar dataset, comprising 44,481 images across 281 vehicle models. The system used unsupervised learning to extract features, followed by training 50 one-vs-rest support vector machines (SVMs) and applying a multi-layer perceptron to estimate confidence scores. If the prediction confidence was below a predefined threshold, the system would output “unknown” as the final class. While deep learning methods achieved slightly higher accuracy, their method reported 97.50\% and 98.40\% accuracy on the IROV and CompuCar datasets, respectively, with less computational overhead, making it suitable for real-time applications.

Although high performance has been reported in traditional image classification tasks, our initial investigations revealed significantly lower accuracy when applying these methods to our real-world dataset (MANPR). Unlike curated benchmark datasets such as Stanford Cars, CompCars, or IROV, where the vehicle of interest (VOI) is typically centred, unobstructed, and dominates the frame, the MANPR dataset contains images where vehicles are often partially visible, surrounded by multiple distractors, or occluded by environmental clutter (e.g., road signs, pedestrians, or other cars). This makes it difficult for standard image classification models, which assign a single label per image, to reliably infer vehicle attributes such as make, shape, or colour. Moreover, many benchmark datasets feature high-resolution, well-lit images captured under controlled conditions, whereas the MANPR dataset consists of lower-quality images captured in real-time under uncontrolled and variable conditions. These differences suggest that while image classification methods perform well under ideal conditions, they do not generalise robustly to complex multi-object scenes typical of law enforcement or traffic surveillance contexts. 

Given the limitations of single-label image classification in complex, real-world environments, we reformulate the task of vehicle metadata recognition as an object detection problem. Object detection models, typically powered by neural networks, not only classify objects but also localise them within an image by predicting bounding boxes around each instance \cite{pedro}. A bounding box refers to a rectangular region that highlights the position of an object within the frame. By providing both class and location information, object detection enables the model to focus on relevant regions, even when multiple objects are present or the target vehicle is partially occluded.

By attending to specific spatial regions within an image, object detection models resemble the way human visual attention prioritises relevant information in cluttered scenes\cite{kubilius2019brainlike, dicarlo2012brain}. This alignment with human perception makes object detection a promising approach for managing the complexity of real-world data, such as the MANPR dataset, where vehicles are often off-centre, partially visible, or captured under variable and uncontrolled conditions. Drawing on principles from the human visual system, which has evolved to constantly operate in dynamic and complex environments, may enhance its robustness in practical AI applications. We therefore hypothesise that object detection models will offer more accurate and generalisable performance for vehicle attribute recognition in real-world scenarios.

Object detection methods can be broadly categorised into two types: two-stage detectors (e.g. Faster Region-Based Convolutional Neural Networks, or Faster R-CNN) and single-stage detectors (e.g. You Only Look Once (YOLO) and Single Shot Detector (SSD))~\cite{ren2015faster, liu2016ssd}. Two-stage detectors perform object detection in two steps: region proposal followed by object classification. Faster R-CNN is a widely used two-stage model that builds upon earlier architectures such as R-CNN and Fast R-CNN by integrating a more efficient region proposal process~\cite{girshick2015fast, ren2015faster}.In Fast R-CNN, region proposals are generated using a selective search, which identifies potential object regions for further processing ~\cite{uijlings2013selective}. Each proposed region is then passed through a CNN to extract features, which are classified using an SVM. Although this approach improved computational efficiency over R-CNN, it still relied on an external slow region proposal method and required separate training for each component.

Faster R-CNN addressed these issues by introducing a region proposal network (RPN) that generates object proposals directly within the CNN pipeline. This integration removed the dependency on selective search and streamlined the model. However, despite these improvements, Faster R-CNN remains a two-stage detector, region proposals are still generated and then passed to a second stage for classification and bounding box refinement. While this improves accuracy, it introduces computational overhead that limits real-time performance.

Traditional object detectors, such as Faster R-CNN, use a two-stage approach: first proposing potential object regions and then classifying them. This mimics a serial attention mechanism in psychology, where attention shifts from one region to another \cite{buschman2009serial}. While accurate, this method is computationally intensive and not ideal for real-time scenarios.

The limitations of two-stage detectors led to the development of single-stage approaches such as YOLO and SSD, which eliminate the region proposal stage altogether\cite{redmon2016you,liu2016ssd}. By streamlining detection into a single pass through the network, these models offer significantly faster inference times, making them well-suited for real-time applications such as traffic surveillance and law enforcement, where rapid and efficient processing is essential. This is particularly relevant for the MANPR dataset, which consists of high-volume, real-time vehicle imagery.

Among these single-stage models, SSD is optimised for detecting small or multi-scale objects by leveraging multiple feature maps from different network layers\cite{liu2016ssd}. While this makes SSD effective in cluttered scenes with varied object sizes, the vehicles in the MANPR dataset generally occupy a substantial portion of the frame. For this reason, we focus on the YOLO family of models, which offer a unified architecture, high-speed performance, and continued accuracy improvements in recent iterations.

Building on the limitations observed with traditional image classification and the demands of real-world detection, we propose using YOLO-based models to extract vehicle metadata from the MANPR dataset. This work aims to establish a baseline for applying object detection techniques over the MANPR dataset. The remainder of this paper is structured as follows. Section 2 outlines the variants of the YOLO models, the structure of the dataset, and our proposed methodology. Section 3 presents the experimental setup, results, and discussion. Section 4 concludes with a summary of the findings and future directions.

\section{Materials and Methods} 
\subsection{You Only Look Once (YOLO)}
YOLO is a single-stage object detection architecture that simultaneously predicts bounding boxes and class probabilities in a single forward pass. By treating detection as a regression problem over spatially divided grid cells, YOLO eliminates the need for a separate region proposal stage, resulting in significant speed improvements compared to two-stage detectors like Faster R-CNN. Because it processes the entire image at once, YOLO also captures contextual information that helps reduce false positives from background clutter.

This design makes YOLO particularly well-suited for real-time applications such as autonomous driving, surveillance, and traffic analysis. Although early versions of YOLO prioritised speed over accuracy, subsequent iterations, such as YOLOv3 and beyond, have introduced substantial accuracy gains without compromising real-time performance~\cite{redmon2018yolov3}.

In this study, we apply YOLO to extract metadata from vehicle images in the MANPR dataset. Its unified framework, real-time inference speed, and consistent performance improvements across versions make it an ideal candidate to benchmark vehicle attribute recognition under real-world conditions. The modular design of the model and its widespread adoption further support its suitability for scalable and latency-sensitive applications such as automated enforcement analytics.

\subsubsection{Detection Process}
Unlike traditional object detection models that analyse images in multiple stages, YOLO processes the entire image in a single forward pass, resulting in significantly more efficient detection~\cite{redmon2016you}. The input image is divided into an $S \times S$ grid, where each grid cell is responsible for predicting bounding boxes for objects whose centres fall within its boundaries. Each cell predicts multiple bounding boxes, along with associated confidence scores that reflect both the presence of an object and the accuracy of the predicted bounding box coordinates. This is done by leveraging a CNN to extract hierarchical feature representations from the image~\cite{redmon2016you, ren2015faster}. The early layers of the CNN capture basic visual patterns (e.g., edges, textures), while deeper layers capture more abstract semantic information. These feature maps are passed through fully connected layers that generate final predictions for bounding boxes and class labels.

\subsubsection{Training Process}
YOLO is trained using a combination of labelled images and penalisation mechanisms. It uses a custom loss function composed of multiple components, each of which penalises specific types of prediction errors (when the prediction deviates from ground truth), such as inaccurate bounding boxes, incorrect class labels or false object presence ~\cite{redmon2016you, redmon2018yolov3}. The following highlights different components within the YOLO training:
\begin{itemize}
    \item \textit{Ground Truth and Penalisation}: During training, the network is provided with annotated datasets containing object classes and bounding box coordinates. The model learns to predict object locations by minimising the difference between predicted bounding boxes and ground truth labels. Incorrect predictions are penalised during training, encouraging the network to iteratively refine its localisation accuracy.

    \item \textit{Loss Function}: YOLO uses a composite loss function that includes three primary components: classification loss, localisation loss, and confidence loss. The localisation component penalises inaccurate bounding box coordinates, the classification component penalises incorrect class predictions, and the confidence component penalises incorrect object presence predictions. This structure allows the model to balance precision in both localisation and classification tasks.

    \item \textit{Class Prediction}: Each grid cell outputs class probabilities that indicate the likelihood of an object belonging to a particular category. These probabilities are refined over multiple training epochs, enabling YOLO to generalise well across diverse object types and backgrounds. YOLO generates multiple class predictions per image, but only the prediction with the highest confidence score, commonly referred to as the top-1 prediction is selected as the final output. This ensures that the model prioritises the most likely classification while discarding less probable alternatives.

\end{itemize}

Through this integrated training process, it learns to perform fast, accurate, and scalable object detection suitable for a range of real-world environments.

\subsubsection{Psychological Basis for How YOLO Works} 
The efficiency of YOLO as an object detection model can be better appreciated through comparisons with human visual processing. YOLO performs detection in a single forward pass, analysing the entire image at once. This mirrors the parallel processing capabilities of the human visual system, where low-level visual features such as colour, orientation, and edges are extracted simultaneously across the visual field \cite{fahle1994human}. According to feature integration theory, early perception involves rapid, parallel processing of basic features, which are then integrated into coherent objects through focused attention. Similarly, YOLO’s convolutional layers perform parallel feature extraction across the image, followed by unified object prediction in a single inference step \cite{treisman1980feature}.

YOLO’s grid-based detection structure reflects Gestalt principles of perceptual organisation, particularly the Law of Proximity and Law of Closure, which describe how the human visual system groups nearby visual elements and fills in missing parts to perceive complete objects\cite{wertheimer1938gestalt}. By dividing the image into a grid, where each cell is responsible for detecting objects within its region, YOLO groups spatially close visual features, echoing the Law of Proximity. It can also recognise partially occluded objects, mirroring the Law of Closure. These mechanisms enable YOLO to interpret cluttered scenes in a way that resembles how the visual system organises incomplete or complex input into coherent forms.

Furthermore, YOLO employs non-maximum suppression (NMS) to resolve overlapping bounding box predictions and retain only the most confident detection. This mechanism reflects the role of selective attention in human cognition, where competing stimuli are filtered to prioritise the most reliable information whilst ignoring redundant inputs \cite{neill1987selective}. 

These architectural choices, parallel processing, spatial structuring, and selective filtering align with core principles of human visual cognition, enabling YOLO to balance speed and accuracy in real-time scenarios such as law enforcement and surveillance.

Given this conceptual alignment, it is intuitive and methodologically sound to extend these psychological frameworks to other aspects of our study. In particular, we apply this lens not only to model selection but also to dataset design. Just as the brain tailors its learning by attending to different types of evidence depending on the task such as fine-grained visual detail when learning vehicle make, or broader contextual cues when learning about colour or shape, we constructed task-specific datasets that reflect these distinct learning demands. This approach ensures that each model is exposed to the most relevant information for its objective, much like how the brain adapts its learning strategy based on the nature of the problem it is trying to solve.

\subsection{YOLO Models}
The aforementioned foundational design makes YOLO an ideal candidate for real-time vehicle metadata recognition tasks. The following section details three variants of the YOLO models: YOLO-v11, YOLO-World, and YOLO-Classification.

\subsubsection{YOLO-v11}
This study utilises YOLO-v11, the latest release in the YOLO object detection family at the time of experimentation \cite{yolo11_ultralytics}. YOLO-v11 introduces several architectural improvements aimed at enhancing detection accuracy and computational efficiency compared to its predecessors. These enhancements include refined feature extraction modules, optimised anchor box strategies, and improved training paradigms. Given its balance of speed and precision, YOLO-v11 was selected for benchmarking across vehicle attribute detection tasks.YOLO-v11 has already been trained on vehicles as a part of the Common Objects in Context (COCO) dataset, with classes highlighting cars, buses, motorcycles, and trucks. 

\subsubsection{YOLO-World}
YOLO-World is a cutting-edge method that enhances the YOLO series of object detectors with open-vocabulary detection capabilities \cite{cheng2024yolo}. In traditional object detection, models are trained to recognise a fixed set of predefined categories. In contrast, an open-vocabulary detector is capable of recognising a broader and potentially unlimited range of object categories, beyond those seen during training, by leveraging textual labels and semantic embeddings. This flexibility enables the model to generalise to novel object classes at inference time. This leads to the concept of zero-shot detection, where a model is able to detect and classify objects that it has never encountered during training. Zero-shot detection is achieved by aligning visual features with language representations, allowing the model to make predictions based on textual descriptions of new classes rather than relying on direct training examples.

YOLO-World achieves this by incorporating a Re-Parameterisable Vision-Language Path Aggregation Network (RepVL-PAN) and a region-text contrastive loss. These components facilitate effective interaction between visual features and linguistic information, enabling the model to associate object regions with their corresponding textual labels, even in the absence of direct training data. Combined with large-scale pre-training on vision-language datasets, this architecture enables YOLO-World to perform robust, flexible object detection across an open and extensible category space.

Note that YOLO World is built on YOLOv8 by incorporating RepVL-PAN to improve feature extraction and fusion. The model is trained on the Common Objects in Context (COCO), which has 80 object classes and Large Vocabulary Instance Segmentation (LVIS) which has 1203 classes with 80 classes in common \cite{COCO,lvis}. These large datasets allows the model to learn a diverse set of classes and their corresponding textual description. In contrast to traditional object detectors that uses a fixed vocabulary and previous open-vocabulary detectors like Adaptive Training Sample Selection (ATSS)\cite{ATSS} and DETR with Improved DeNoising Anchor Boxes (DINO)\cite{DINO}, which has high computational and deployment requirements, YOLO-World uses a light weight YOLOv8 object detector and an offline vocabulary to encode the user prompts thus enabling real-time inference and easier deployment.

The text embeddings from the user prompts are extracted using the transformer text encoder pre-trained by Contrastive Language-Image Pre-training (CLIP) \cite{CLIP}. During training, an online vocabulary is constructed but during inference the embeddings for the prompts are generated from an offline vocabulary, which reduces the computational costs and allows to easily adjust the vocabulary. The prompts from the user can be a category name or a noun phrase or an object description. During pre-training rather than using image-text pairs an automatic labelling technique is used to generate region-text pairs. First, the noun phrases from the text are extracted and using Grounded Language-Image Pre-training (GLIP)  pseudo boxes are generated for the noun phrases \cite{GLIP}. These coarse region-text pairs are then filtered to remove low relevance pairs. Employing the aforementioned techniques the YOLO-World model is shown to be superior in terms of open-vocabulary performance and speed.

\subsubsection{YOLO-Classification}
While most previous vehicle metadata classification approaches rely on conventional CNN-based image classifiers, recent developments have introduced unified architectures that extend beyond traditional classification models. With the YOLO family of models, which are primarily designed for real-time object detection, variants have been implemented to support classification \cite{vstancel2019introduction, jiang2022review}. YOLO-Classification re-purposes the YOLO architecture originally designed for detecting and localising multiple objects into a high-speed, single-label image classification pipeline. It benefits from the same underlying architecture used in object detection, offering a consistent framework for both detection and classification tasks. YOLO-Classification has the advantage of being computationally efficient and highly scalable, making it particularly attractive for real-time or resource-constrained deployments.

In our study, we incorporated YOLO-Classification as a baseline image classification method to evaluate its performance on vehicle metadata tasks, including make,shape, and colour recognition. This approach allows for a direct comparison with YOLO's detection counterparts (e.g., YOLO-v11 and YOLO-World), enabling us to assess whether the classification-only variant is sufficient for identifying vehicle attributes or whether localisation-aware models offer performance benefits. Given its speed, simplicity, and compatibility with other YOLO-based models, YOLO classification provides a practical and consistent starting point for benchmarking across different metadata prediction tasks.

\subsection{YOLO-v11 vs YOLO-World vs YOLO-Classification}
Given their complementary strengths, this study aims to compare YOLO-v11, YOLO-World, and YOLO-Classification to evaluate their effectiveness in vehicle metadata recognition tasks. YOLO-v11 represents the latest advancement in the traditional YOLO pipeline, optimised for high-speed, closed-set object detection with strong performance on predefined categories. In contrast, YOLO-World extends this framework by introducing open-vocabulary detection capabilities through vision-language modelling, enabling the identification of previously unseen classes in a zero-shot setting. This comparison allows us to assess the trade-offs between closed-set accuracy and open-set flexibility, an important consideration in real-world policing scenarios where class definitions may be incomplete, ambiguous, or subject to change.

In addition to detection models YOLO-v11 and YOLO-World, we include YOLO-Classification in our comparison. While YOLO is primarily known for object detection, its architecture has been adapted for image classification tasks by modifying the output layers to produce class probabilities instead of bounding boxes. This is particularly effective when the object of interest dominates the image, as is the case with many vehicle images in the MANPR dataset, making classification a lightweight and computationally efficient alternative. Comparing YOLO Classification with detection-based approaches allows us to determine whether explicit localisation provides a significant performance benefit in the context of vehicle metadata recognition.

\subsection{Multi-View Inference (MVI) with Majority Voting}
This study focuses on evaluating the capabilities of existing state-of-the-art AI techniques and determining which is best suited for practical deployment within a policing context. Given the operational risks associated with misclassification, high model accuracy is a critical requirement. At the same time, real-world deployments must also consider speed and memory efficiency, especially for systems intended to run on resource-constrained devices or in real-time settings. Although many existing approaches focus primarily on improving detection accuracy, relatively fewer studies offer a comparative analysis of the trade-offs between accuracy and inference speed, factors that are crucial for deployment in mission-critical environments such as policing. Therefore, our evaluation focuses on two key criteria: classification accuracy and inference speed.

Understanding the psychological foundations of learning informs how we structure training data. As mentioned earlier, just as humans rely on different types of information to learn different object properties, our models should be trained in a task-specific manner. From a cognitive science perspective, learning is shaped by the relevance of cues and feedback associated with each task~\cite{friston2005theory, richardson2006memory}. For example, identifying a vehicle’s make often depends on fine-grained visual features such as emblems and body contours. In contrast, recognising colour relies on broader surface-level cues and is influenced by lighting and context. Shape recognition lies somewhere in between, requiring attention to structural outlines and proportions.

Using a single dataset to train all models assumes that the same visual input is equally informative across these distinct recognition tasks, an assumption that contradicts both psychological theories of task-specific encoding and machine learning principles of specialised feature learning ~\cite{tyler2013objects}. To address this, we curated separate training and validation datasets for each task (colour, make, and shape) ensuring that each model was optimised to focus on the most relevant visual cues for its prediction objective.

Importantly, we retained a standardised test dataset across all models to enable fair performance comparison. This ensures that any observed differences reflect the effectiveness of task-specific training, rather than disparities in test conditions, thus preserving experimental control.

\begin{figure}[!ht]
\centering
\includegraphics[width=\linewidth]{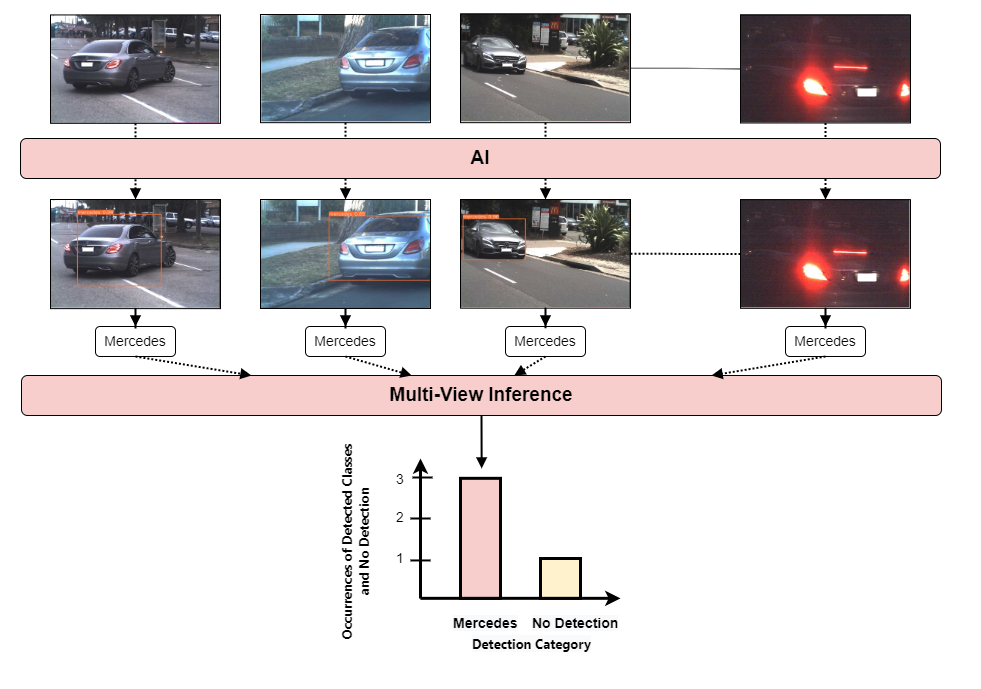}
\caption{Majority voting across multiple images of the same vehicle: Each image is independently analysed by the AI model to predict make. The predictions are aggregated, and the most frequently predicted class is selected as the final label. In this example, three out of four images, the AI predicts "Mercedes" three times and "No Detection" one time, resulting in a final classification of the vehicle as a "Mercedes".}
\label{fig:multiview}
\end{figure}

To further enhance robustness under real-world variability, we implemented a multi-view inference strategy. Humans rarely make identity or attribute judgements from a single viewpoint; instead, we integrate information across multiple perspectives to form more reliable conclusions~\cite{brojde2012bilingual}. Mimicking this process, our approach generates predictions for each vehicle across multiple images, captured from varying angles or lighting conditions across different time frames, and applies majority voting to determine the final label. This ensemble-style method improves resilience to occlusion, pose variation, and resolution differences. 
\autoref{fig:multiview} illustrates this inference strategy. It is worth mentioning that such an approach was mainly proposed to enhance searching and filtering of a set of large vehicles images. 

YOLO models are available in different sizes namely, nano, small, medium, large and x-large and are used to balance trade-off between speed, computational efficiency and accuracy. The smaller models are used on edge devices with limited resources and are optimised for real-time performance, while the larger models provide higher accuracy where computational power is less constrained. These variants differ in network depth, trainable parameter count and width, thus allowing users to select a model that best fits their requirements. We utilised three variants within each of the three selected models (small, large, and x-large).  This comparison provides a useful baseline for understanding trade-offs between accuracy and computational cost across different model scales.

\subsection{Mobile Automated Number Plate Recognition (MANPR) Data} 
MANPR plays a crucial role in vehicle identification for law enforcement and traffic monitoring \cite{anpr_survey}. Many existing datasets are captured in controlled environments, such as dealerships or fixed surveillance points, where images are taken under ideal conditions. While these datasets are useful, they do not fully represent real-world challenges. In contrast, this dataset was collected in dynamic, real-world scenarios, where various external factors affect image quality and the ability to recognise number plates accurately. Examples of these external factors include: 
\begin{itemize}
    \item Distance between the capturing camera and the vehicle, affecting resolution and clarity.
    \item Motion blur caused by the movement of the police vehicle relative to the detected vehicle.
    \item Lighting conditions, reflections, and glare, which can obscure number plate details.
    \item Weather conditions such as rain, fog, or dust, which may distort visibility.
    \item The presence of multiple vehicles in a single frame, leading to occlusions or overlapping objects.
\end{itemize}

The MANPR dataset was collected across various times of the day, including both daylight and night conditions. The images also include multiple viewpoints of vehicles, such as front, front-side, rear, rear-side, and side angles. Additionally, multiple images of the same vehicle were captured under different conditions, creating a more complex and representative dataset compared to those used in previous studies.

\subsubsection{Labelling Process for Vehicle Attributes of MANPR Dataset}
To incorporate make, shape, and colour labels, a structured annotation process was followed. A labelling tool was developed using Streamlit and Python, and  annotators were recruited to label the images that belong to a number plate. The labelling process involved:
\begin{enumerate}
    \item Grouping images by number plate to ensure that annotators could view multiple images of the same vehicle from different perspectives (front, rear, side).
    \item Verifying the number plate to ensure accurate data association before proceeding with further labelling.
    \item Identifying vehicle make based on their existing knowledge of the manufacturers. This step required annotators to compare images with reference models when necessary.
    \item Classifying vehicle shape into broad categories such as sedan, SUV, hatchback, van, or truck.
    \item Determining vehicle colour based on visible characteristics in the image. While lighting conditions can affect colour perception, annotators followed guidelines to standardise labelling.
\end{enumerate}

Note that most images were collected between 6 AM and 8 PM to maintain optimal lighting conditions. However, night time images were not entirely excluded, ensuring the dataset remains representative of real-world scenarios and that the system can operate under varied lighting conditions. A different annotator would then review the labelled vehicle annotations to confirm their accuracy. If the annotations were correct, they would be left unchanged. If errors were found, the annotator would update the labels to the correct one.
  
\subsubsection{Data Cleaning and Filtering}
Given the complexity of real-world image capture, a multi-stage filtering process was applied to ensure dataset quality. YOLO is well-established as an effective model for detecting vehicles in images. Accordingly, we used a YOLO-based object detection model to localise vehicles within our dataset.\cite{yolov8}. Since YOLO often outputs multiple bounding boxes, a filtering strategy was implemented to retain only the most relevant detection.
\begin{enumerate}
    \item \textit{Bounding box selection: }The area of each bounding box was calculated (width × height), and only the first prediction was retained if it had the largest bounding box. This approach assumes that the object with the largest bounding box corresponds to the largest, and therefore most prominent, object in the image making it the most relevant detection.
    \item \textit{Ensuring unique number plates:} Some number plates appeared across multiple vehicles due to individuals in NSW using the same number plate when purchasing a new or different car. These records were cross-checked, and cases where a single plate was linked to different vehicles were removed. 
    \item \textit{Merging similar classes:} Beyond issues of imbalance, early visual inspection also uncovered inconsistencies and semantic redundancies in the labelling process. For instance, in the colour data, closely related shades like Red and Maroon, or Beige and Gold, were separated into distinct labels despite often being indistinguishable under real-world lighting conditions.

In addition to removing low-frequency classes, we also consolidated semantically redundant labels. For example, Red and Maroon were merged because of the high visual similarity, as were Beige and Gold. These refinements ensured clearer label semantics, reduced class confusion, and improved interpretability of the models.

\end{enumerate}

This multi-stage data cleaning and restructuring process ensured that our datasets were well-aligned with the learning requirements of each task. By combining statistical filtering with psychological and perceptual insights, we enhanced the dataset’s clarity, reduced label noise, and maximised the models’ ability to learn discriminative features efficiently.

\subsubsection{Task-Specific Datasets for Make, Shape, Colour, and Colour-binary}
For each of the attributes, the metadata extraction task depends on distinct types of visual evidence: extracting the make requires fine-grained logo and styling features, extracting the colour relies on pixel-level shading, and the shape identification is tied to overall silhouette. To support optimal learning, we curated separate datasets for each task, tailored to their unique learning goals. We introduced a second version of the colour dataset, categorising colours into bright and dark groups to account for variations in lighting, hence forth referred to as colour-binary. All datasets were split using the same principles:
\begin{itemize}
    \item 30\% was reserved for testing.
    \item The remaining 70\% was split 80/20 for training and validation.
    \item No number plate appeared in more than one dataset to prevent data leakage.
\end{itemize}

We maintained a consistent test dataset across all the model types (make, shape, colour, and colour-binary). Importantly, all filters and class modifications applied during training, such as the removal of rare classes, consolidation of semantically similar labels (e.g., merging "Red" and "Maroon"), and exclusion of ambiguous categories were also applied to the test dataset. This ensured that the test dataset did not introduce any classes or labels that were absent during training, preserving the integrity of model evaluation and preventing unrealistic generalisation to unseen categories. See \autoref{tab:dataset_summary} for a break down of the number of images available in the train, validation, and test sets for the make, shape, colour, and colour-binary  datasets.

Each image was labelled with ground truth annotations comprising bounding box coordinates and corresponding class information. These were stored in YOLO’s standard format, with each image accompanied by a text file detailing the object’s position and its class label. The directory was organised according to YOLO’s expected structure.

\begin{table}[h!]
\centering
\caption{Summary of image and number plate counts across the train, validation, and test datasets.}
\begin{tabular}{ccc|cc|cc|cc}
\hline
\textbf{Dataset} & \textbf{Total Classes} & \textbf{Total Plates} & 
\multicolumn{2}{c|}{\textbf{Train}} & 
\multicolumn{2}{c|}{\textbf{Validation}} & 
\multicolumn{2}{c}{\textbf{Test}} \\
\cline{4-9}
& & & \textbf{Images} & \textbf{Plates} & \textbf{Images} & \textbf{Plates} & \textbf{Images} & \textbf{Plates} \\
\hline
make & 31 & 6,945 & 68,041 & 4,108 & 17,056 & 1,028 & 29,937 & 1,809 \\
shape & 6 & 6,857 & 67,152 & 4,038 & 16,777 & 1,010 & 29,937 & 1,809 \\
colour & 9 & 6,937 & 67,938 & 4,102 & 16,691 & 1,026 & 29,937 & 1,809 \\
colour-binary & 2 & 6,937 & 67,938 & 4,102 & 16,691 & 1,026 & 29,937 & 1,809 \\
\hline
\end{tabular}
\label{tab:dataset_summary}
\end{table}

\section{Experiments and Results} 

\subsection{Experimental Setup}
A Nitro GN29A Gen2 2U NVLink GPU Server was utilised to conduct the experiments. The server contained 4 NVIDIA A100 GPUs with 40GB each. Openshift AI, which is a product from Redhat, was used to conduct the experiments. For each task (make, shape, colour, and colour-binary), three models were selected (YOLO-v11, YOLO-World, YOLO-Classification). With each model, three variants were included (small, large, x-large), resulting in a total of 36 experiments. Each model was trained with a batch size of 8 for 30 epochs, using an image resolution of 640×640. Stochastic Gradient Descent (SGD) was employed as the optimisation algorithm, with a learning rate set to 0.01.

\subsection{Evaluation}

This study focused on evaluating the performance of different models in extracting vehicle metadata, such as colour, make, and shape, from images. The intention was not to re-assess their general object detection capabilities, as it has already been well studied and reported in literature.

The test dataset consisted of 29,937 images, belonging to 1,809 number plates. 
Two evaluation strategies were introduced. The first refers to a multi-view inference (MVI) which is proposed in our approach. The second is referred to as single-view inference (SVI). With MVI, the information extracted from each image is grouped according to the corresponding number plate. Then, majority voting is applied before calculating the classification accuracy of the model. With SVI, each image is treated as a separate entity.

\subsection{Results and Analyses}

\subsubsection{Performance Across Vehicle Attribute Models
}The results reporting the performance for each of the vehicle attribute models across each dataset are shown in Tables~\ref{tab:make_acc},~\ref{tab:shape_acc},~\ref{tab:colour_acc},~\ref{tab:colour-two_acc} respectively. As shown in \autoref{tab:make_acc}, to identify the make of vehicles, the YOLO-v11 model demonstrated strong performance, with classification accuracy ranging from 86.43\% to 88.16\% with SVI and 92.87\% to 93.31\% with MVI. The YOLO-World model had similar results, with classification accuracy between 86.02\% and 88.04\% with SVI and 93.26\% to 93.70\% with MVI. The YOLO-Classification model showed a slight decrease in classification accuracy, with scores ranging from 80.41\% to 84.04\% with SVI and 91.43\% to 92.43\% with MVI. 

When it comes to identifying the shape of vehicles, the YOLO-v11 and YOLO-World models demonstrated comparable performance with MVI. The YOLO-Classification model however showed a slight decrease in classification accuracy, with scores ranging from 74.97\% to 75.88\% with SVI and 80.04\% to 80.87\% with MVI.

For identifying the colour of vehicles, the YOLO-v11 model achieved classification accuracy ranging from 78.68\% to 79.24\% with SVI and 84.47\% to 84.74\% MVI. The YOLO-World model showed similar performance, with classification accuracy between 77.84\% and 78.54\% with SVI and 84.02\% to 84.13\% with MVI. Interestingly, the YOLO-Classification approach had better classification accuracy, with the best model reporting 85.19\% with MVI. 

\autoref{tab:colour-two_acc} reports the models performance when extracting bright and dark colours. The YOLO-v11 model achieved classification accuracy ranging from 91.98\% to 92.01\% SVI and 94.25\% to 94.75\% MVI. The YOLO-World model showed similar performance, with classification accuracy between 91.66\% and 91.83\% SVI and 94.36\% to 94.80\% MVI. The YOLO-Classification model had classification accuracy ranging from 91.76\% to 91.96\% SVI and 94.64\% to 94.86\% MVI.

\begin{table}[h!]
\centering
\caption{Comparative performance of models in extracting the make attribute from vehicles.}
\begin{tabular}{c|c|c|c|c}
\hline
\textbf{Model} & \textbf{Inference} & \textbf{Small} & \textbf{Large} & \textbf{X-Large}  \\
\hline
\multirow{2}{*}{YOLO-v11} & SVI  & 86.43 & 87.59 & 88.16 \\& MVI & 92.87 & 93.42 & 93.31 \\       
\hline
\multirow{2}{*}{YOLO-World}& SVI  & 86.02 & 87.74 & 88.04 \\  & MVI & \textbf{93.70} & 93.42 & 93.26 \\               
\hline
\multirow{2}{*}{YOLO-Classification}& SVI  & 80.41 & 83.39 & 84.04 \\  & MVI & 91.43 & 91.87 & 92.43 \\
                         
\hline
\end{tabular}
\label{tab:make_acc}
\end{table}

\begin{table}[h!]
\centering
\caption{Comparative performance of models in extracting the shape attribute from vehicles.}
\begin{tabular}{c|c|c|c|c}
\hline
\textbf{Model} & \textbf{Inference} & \textbf{Small} & \textbf{Large} & \textbf{X-Large}  \\
\hline
\multirow{2}{*}{YOLO-v11} & SVI & 78.43 & 78.98 & 78.94 \\& MVI & 82.26 & 82.31 & \textbf{82.86} \\
 
\hline
\multirow{2}{*}{YOLO-World}& SVI & 78.04 & 77.62 & 77.37 \\ & MVI & 82.81 & 81.98 & 82.70 \\
 
\hline
\multirow{2}{*}{YOLO-Classification}& SVI & 74.97 & 75.88 & 75.01 \\ & MVI & 80.04 & 80.87 & 80.15 \\
 
\hline
\end{tabular}
\label{tab:shape_acc}
\end{table}

\begin{table}[h!]
\centering
\caption{Comparative performance of models in extracting the colour attribute from vehicles.}
\begin{tabular}{c|c|c|c|c}
\hline
\textbf{Model} & \textbf{Inference} & \textbf{Small} & \textbf{Large} & \textbf{X-Large}  \\
\hline
\multirow{2}{*}{YOLO-v11} & SVI& 78.68 & 79.35 & 79.24 \\
& MVI & 84.74 & 84.69 & 84.47 \\

\hline
\multirow{2}{*}{YOLO-World} & SVI& 77.84 & 78.32 & 78.54 \\ & MVI & 84.08 & 84.02 & 84.13 \\

\hline
\multirow{2}{*}{YOLO-Classification}  & SVI & 77.41 & 78.06 & 78.06 \\& MVI & 84.58 & \textbf{85.19} & 84.96 \\

\hline
\end{tabular}
\label{tab:colour_acc}
\end{table}

\begin{table}[h!]
\centering
\caption{Comparative performance of models in extracting the colour (bright/dark) attributes from vehicles.}
\begin{tabular}{c|c|c|c|c}
\hline
\textbf{Model} & \textbf{Inference} & \textbf{Small} & \textbf{Large} & \textbf{X-Large}  \\
\hline
\multirow{2}{*}{YOLO-v11} & SVI& 91.98 & 92.01 & 91.99 \\
& MVI & 94.75 & 94.25 & 94.53 \\

\hline
\multirow{2}{*}{YOLO-World} & SVI& 91.66 & 91.83 & 91.71 \\ & MVI & 94.80 & 94.64 & 94.36 \\

\hline
\multirow{2}{*}{YOLO-Classification}  & SVI & 91.76 & 91.92 & 91.96 \\& MVI & 94.80 & \textbf{94.86} & 94.64 \\

\hline
\end{tabular}
\label{tab:colour-two_acc}
\end{table}

The experiments involved training four distinct models, each specialising in recognising either the make, shape, colour, or bright/dark colours of vehicles in images. The accuracy of these models was measured in two ways: SVI and MVI. The MVI approach improved the performance compared to SVI in all of the experiments. Since there are multiple images associated with each number plate, the overall MVI accuracy tends to be higher than the SVI accuracy. This discrepancy can be attributed to the variability in image quality; some images may clearly display the vehicle's colour, make, and shape, while others may not. By selecting the best response from the set of images associated with each number plate, the models can achieve better accuracy metrics. We conclude that using MVI is essential in our scenario, as it consistently outperforms SVI due to its ability to leverage multiple views per vehicle. This is feasible given the high accuracy of the number plate detection system, which allows reliable grouping of related images.

\subsubsection{Comparing accuracy across YOLO types (Detection v Classification)}
Overall, YOLO detection models outperformed classification models on the make and shape tasks, while classification models performed slightly better on the colour and binary-colour tasks. The binary-colour models achieved the highest performance across all tasks. This is likely due to the clear visual separation between bright and dark classes, which made it easier for models to learn distinctive features. The highest reported accuracy was 94.86\% (YOLO-Classification, MVI), highlighting the benefit of reduced intra-class variability, leading to better discriminative features among classes.

The make models were the second-best performing group. As expected, vehicle make classification benefited from the presence of identifiable visual elements such as logos and grille patterns. The best-performing model achieved 93.70\% accuracy across 31 classes, demonstrating high precision in recognising fine-grained brand-level differences.

The colour models ranked third, with the top YOLO-Classification model achieving 85.19\% accuracy using MVI. While still useful for filtering and searching vehicle images, colour classification showed a noticeable drop in performance compared to binary-colour. This may be due to the increased difficulty of distinguishing between visually similar colours (e.g., silver vs grey) under varying lighting conditions.

The shape models had the lowest performance overall, with a best accuracy of 82.86\%. This may reflect the high visual similarity across modern vehicle designs. For example, hatchbacks and compact SUVs often share similar silhouettes, making it harder to reliably distinguish between shape categories. Nevertheless, shape classification can still support search and filtering pipelines. Based on our results, shape-based filtering would reduce search time in approximately 82.86\% of queries, aligning with our objective of speeding up identification of VOIs.

These results suggest that tasks with more visually distinct classes (e.g., binary colour) yield higher accuracy, while tasks involving more subtle distinctions (e.g., shape) pose greater challenges. Importantly, for both the make and shape tasks, detection models consistently outperformed classification models across all experiments. This underscores the value of using object detection approaches for metadata extraction in complex, real-world scenarios.

\subsubsection{Comparing accuracy across YOLO versions}
\begin{figure}[!ht]
\centering
\includegraphics[width=\linewidth]{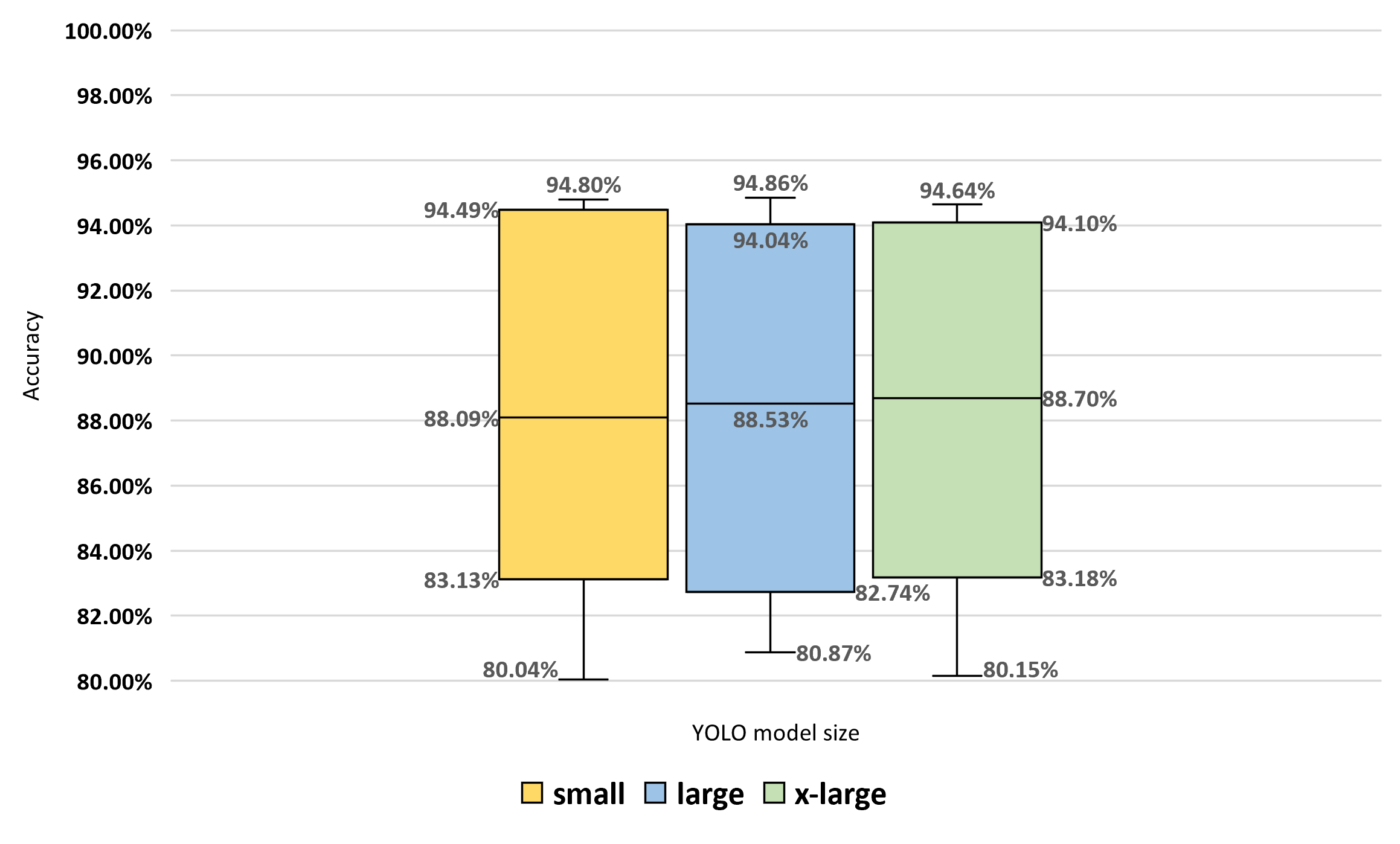}
\caption{Performance comparison of small, large, and x-large models using the MVI approach.}
\label{fig:boxplot}
\end{figure}

Taking into consideration the large number of records to be processed within the policing context, we analysed the effect of using different versions of the detection and classification models. \autoref{fig:boxplot} presents the classification accuracy distributions for each group of models (small, large, and x-large) using the MVI approach. As shown, there is no substantial performance difference between model sizes when extracting vehicle metadata. Notably, the small YOLO-WORLD model reported the best performance with the make dataset (93.70\%). The small YOLO-World model achieved the second-best accuracy on the shape dataset (82.81\%), while the small YOLO-v11 model ranked second on the colour dataset (84.74\%). For the binary-colour task, both the small YOLO-World and YOLO-Classification models reached 94.80\% accuracy.

These small variations across model sizes suggest that compact models are well-suited for large-scale applications, offering nearly equivalent performance with lower computational cost. In fact, increasing model size beyond a certain point did not lead to consistent accuracy gains and, in some cases, slightly reduced performance. Overall, our findings indicate that small models are sufficient and practical for vehicle metadata extraction in real-world scenarios.
\subsubsection{Comparing YOLO-WORLD models with vs without fine tuning}

With the rise of zero-shot learning in vision models, we explored whether off-the-shelf YOLO-World models could be used for vehicle metadata extraction without fine-tuning. We evaluated their performance across the three datasets in this setting. While the best zero-shot results were observed for the shape dataset, the overall performance remained insufficient for direct application to a dataset as complex as MANPR. As shown in \autoref{table:combined}, fine-tuning improved model accuracy dramatically, by up to 90 percentage points in some cases, demonstrating that zero-shot approaches alone are insufficient for complex, real-world datasets like MANPR.
\begin{table}[h!]
\centering
\caption{Classification Accuracy for Make, Shape, and Colour Identification with and without YOLO-World Fine Tuning}
\begin{tabular}{c|c|cc}
\hline
\textbf{Attribute} & \textbf{Size} & \textbf{No Fine Tuning (\%)} & \textbf{Fine Tuning (\%)} \\
\hline
\multirow{3}{*}{Make} & Small & 1.80 & 93.70 \\
 & Large & 2.20 & 93.42 \\
 & X-Large & 2.30 & 93.26\\
\hline
\multirow{3}{*}{Shape} & Small & 36.70 & 82.81 \\
 & Large & 37.00 & 81.98 \\
 & X-Large & 38.40 & 82.70 \\
\hline
\multirow{3}{*}{Colour} & Small & 15.40 & 84.08 \\
 & Large & 0.00 & 84.02 \\
 & X-Large & 1.70 & 84.13 \\
\hline
\end{tabular}
\label{table:combined}
\end{table}

\subsubsection{Handling Uncertainty in Detection: Implications for Policing}
In certain cases, the models generated no detections. While this might appear to be a limitation, it is in fact desirable in policing contexts where a false detection could lead to incorrect matches. In such applications, an 'unknown/no-detection' output is often preferable to a misclassification. This reinforces our preference for detection models, which showed better performance in handling uncertainty, and supports their integration into operational use cases.

\section{Conclusion}
This paper presented a comprehensive evaluation of three YOLO-based models YOLO-v11, YOLO-World, and YOLO-Classification on vehicle metadata extraction tasks, including make, colour, and shape classification. We developed a dataset of over 100,000 real-world vehicle images and employed a multi-view inference approach, where multiple images associated with the same vehicle are used to generate a final prediction. Our experiments demonstrated that detection-based models outperformed classification models on the make and shape tasks, achieving accuracies of up to 94.86\% and 82.86\%, respectively. These findings suggest that detection models are better suited to complex, real-world datasets where object localisation enhances attribute recognition. For the colour and binary-colour tasks, the best performance was achieved by classification models, with accuracies of 94.86\% and 85.19\%, highlighting their effectiveness for visually distinct categories. We also found that smaller models achieved comparable performance to larger variants while offering significant gains in efficiency. This makes them particularly well-suited for resource-constrained environments, such as mobile or in-vehicle deployments in law enforcement applications. Overall, this work provides a strong baseline for understanding the challenges of vehicle metadata extraction in operational settings and for evaluating the capabilities of existing YOLO-based frameworks. Future work will focus on improving dataset quality, refining label structures, and experimenting with other architectures, such as large vision models. We also aim to explore part-based approaches, where models are trained on specific vehicle components and their predictions are aggregated to improve overall accuracy.

\section{Acknowledgements}
The authors wish to acknowledge the support and assistance of the NSWPF in undertaking this research. The views expressed in this publication are not necessarily those of the NSWPF and any errors of omission or commission are the responsibility of the authors. This work was conducted as part of a registered internal research project, identified by D/2024/1063120. This work was conducted as part of an internship supported by the Defence Innovation Network (DIN). Special thanks to Chief Inspector Matthew Mccarthy for his help in enabling the internship, and to Sergeant Tamzin Drakes for recruiting the labellers. We also appreciate the labellers for their diligent efforts in labelling the data for us. We would like to extend our gratitude to Chris Butler, Tom Corcoran, Daniel Cifuentes and David Bain from the Red Hat team, as well as Sharia Chowdhury and Dr Xinguang Wang for their invaluable assistance in setting up the original infrastructure.  

\bibliography{sample}

\begin{thebibliography}{10}
\urlstyle{rm}
\expandafter\ifx\csname url\endcsname\relax
  \def\url#1{\texttt{#1}}\fi
\expandafter\ifx\csname urlprefix\endcsname\relax\def\urlprefix{URL }\fi
\expandafter\ifx\csname doiprefix\endcsname\relax\def\doiprefix{DOI: }\fi
\providecommand{\bibinfo}[2]{#2}
\providecommand{\eprint}[2][]{\url{#2}}

\bibitem{nazemi_real-time_2019}
\bibinfo{author}{Nazemi, A.}, \bibinfo{author}{Azimifar, Z.}, \bibinfo{author}{Shafiee, M.~J.} \& \bibinfo{author}{Wong, A.}
\newblock \bibinfo{journal}{\bibinfo{title}{Real-time vehicle make and model recognition using unsupervised feature learning}}.
\newblock {\emph{\JournalTitle{IEEE Transactions on Intelligent Transportation Systems}}} \textbf{\bibinfo{volume}{21}}, \bibinfo{pages}{3080--3090} (\bibinfo{year}{2019}).
\newblock \bibinfo{note}{ISBN: 1524-9050 Publisher: IEEE}.

\bibitem{tan_artificial_2024}
\bibinfo{author}{Tan, S.~H.}, \bibinfo{author}{Chuah, J.~H.}, \bibinfo{author}{Chow, C.-O.}, \bibinfo{author}{Kanesan, J.} \& \bibinfo{author}{Leong, H.~Y.}
\newblock \bibinfo{journal}{\bibinfo{title}{Artificial intelligent systems for vehicle classification: {A} survey}}.
\newblock {\emph{\JournalTitle{Engineering Applications of Artificial Intelligence}}} \textbf{\bibinfo{volume}{129}}, \bibinfo{pages}{107497}, \doiprefix\url{10.1016/j.engappai.2023.107497} (\bibinfo{year}{2024}).

\bibitem{deshmukh_swin_2023}
\bibinfo{author}{Deshmukh, P.}, \bibinfo{author}{Satyanarayana, G. S.~R.}, \bibinfo{author}{Majhi, S.}, \bibinfo{author}{Sahoo, U.~K.} \& \bibinfo{author}{Das, S.~K.}
\newblock \bibinfo{journal}{\bibinfo{title}{Swin transformer based vehicle detection in undisciplined traffic environment}}.
\newblock {\emph{\JournalTitle{Expert Systems with Applications}}} \textbf{\bibinfo{volume}{213}}, \bibinfo{pages}{118992}, \doiprefix\url{10.1016/j.eswa.2022.118992} (\bibinfo{year}{2023}).

\bibitem{surwase_multi-scale_2023}
\bibinfo{author}{Surwase, S.} \& \bibinfo{author}{Pawar, M.}
\newblock \bibinfo{journal}{\bibinfo{title}{Multi-scale multi-stream deep network for car logo recognition}}.
\newblock {\emph{\JournalTitle{Evolutionary Intelligence}}} \textbf{\bibinfo{volume}{16}}, \bibinfo{pages}{485--492}, \doiprefix\url{10.1007/s12065-021-00671-1} (\bibinfo{year}{2023}).

\bibitem{gayen_two_2024}
\bibinfo{author}{Gayen, S.}, \bibinfo{author}{Maity, S.}, \bibinfo{author}{Singh, P.~K.}, \bibinfo{author}{Geem, Z.~W.} \& \bibinfo{author}{Sarkar, R.}
\newblock \bibinfo{journal}{\bibinfo{title}{Two decades of vehicle make and model recognition – {Survey}, challenges and future directions}}.
\newblock {\emph{\JournalTitle{Journal of King Saud University - Computer and Information Sciences}}} \textbf{\bibinfo{volume}{36}}, \bibinfo{pages}{101885}, \doiprefix\url{10.1016/j.jksuci.2023.101885} (\bibinfo{year}{2024}).

\bibitem{krizhevsky_imagenet_2012}
\bibinfo{author}{Krizhevsky, A.}, \bibinfo{author}{Sutskever, I.} \& \bibinfo{author}{Hinton, G.~E.}
\newblock \bibinfo{title}{{ImageNet} {Classification} with {Deep} {Convolutional} {Neural} {Networks}}.
\newblock In \emph{\bibinfo{booktitle}{Advances in {Neural} {Information} {Processing} {Systems}}}, vol.~\bibinfo{volume}{25} (\bibinfo{publisher}{Curran Associates, Inc.}, \bibinfo{year}{2012}).

\bibitem{ma_ai-based_2020}
\bibinfo{author}{Ma, X.} \& \bibinfo{author}{Boukerche, A.}
\newblock \bibinfo{title}{An ai-based visual attention model for vehicle make and model recognition}.
\newblock In \emph{\bibinfo{booktitle}{2020 {IEEE} {Symposium} on computers and communications ({ISCC})}}, \bibinfo{pages}{1--6} (\bibinfo{publisher}{IEEE}, \bibinfo{year}{2020}).

\bibitem{amirkhani_deepcar_2022}
\bibinfo{author}{Amirkhani, A.} \& \bibinfo{author}{Barshooi, A.~H.}
\newblock \bibinfo{journal}{\bibinfo{title}{{DeepCar} 5.0: vehicle make and model recognition under challenging conditions}}.
\newblock {\emph{\JournalTitle{IEEE Transactions on Intelligent Transportation Systems}}} \textbf{\bibinfo{volume}{24}}, \bibinfo{pages}{541--553} (\bibinfo{year}{2022}).
\newblock \bibinfo{note}{ISBN: 1524-9050 Publisher: IEEE}.

\bibitem{yan_exploiting_2017}
\bibinfo{author}{Yan, K.}, \bibinfo{author}{Tian, Y.}, \bibinfo{author}{Wang, Y.}, \bibinfo{author}{Zeng, W.} \& \bibinfo{author}{Huang, T.}
\newblock \bibinfo{title}{Exploiting multi-grain ranking constraints for precisely searching visually-similar vehicles}.
\newblock In \emph{\bibinfo{booktitle}{Proceedings of the {IEEE} international conference on computer vision}}, \bibinfo{pages}{562--570} (\bibinfo{year}{2017}).

\bibitem{tang_cityflow_2019}
\bibinfo{author}{Tang, Z.} \emph{et~al.}
\newblock \bibinfo{title}{Cityflow: {A} city-scale benchmark for multi-target multi-camera vehicle tracking and re-identification}.
\newblock In \emph{\bibinfo{booktitle}{Proceedings of the {IEEE}/{CVF} {Conference} on {Computer} {Vision} and {Pattern} {Recognition}}}, \bibinfo{pages}{8797--8806} (\bibinfo{year}{2019}).

\bibitem{liu_deep_2016}
\bibinfo{author}{Liu, H.}, \bibinfo{author}{Tian, Y.}, \bibinfo{author}{Yang, Y.}, \bibinfo{author}{Pang, L.} \& \bibinfo{author}{Huang, T.}
\newblock \bibinfo{title}{Deep relative distance learning: {Tell} the difference between similar vehicles}.
\newblock In \emph{\bibinfo{booktitle}{Proceedings of the {IEEE} conference on computer vision and pattern recognition}}, \bibinfo{pages}{2167--2175} (\bibinfo{year}{2016}).

\bibitem{bai_group-sensitive_2018}
\bibinfo{author}{Bai, Y.} \emph{et~al.}
\newblock \bibinfo{journal}{\bibinfo{title}{Group-sensitive triplet embedding for vehicle reidentification}}.
\newblock {\emph{\JournalTitle{IEEE Transactions on Multimedia}}} \textbf{\bibinfo{volume}{20}}, \bibinfo{pages}{2385--2399} (\bibinfo{year}{2018}).
\newblock \bibinfo{note}{ISBN: 1520-9210 Publisher: IEEE}.

\bibitem{liu_deep_2016-1}
\bibinfo{author}{Liu, X.}, \bibinfo{author}{Liu, W.}, \bibinfo{author}{Mei, T.} \& \bibinfo{author}{Ma, H.}
\newblock \bibinfo{title}{A deep learning-based approach to progressive vehicle re-identification for urban surveillance}.
\newblock In \emph{\bibinfo{booktitle}{Computer {Vision}–{ECCV} 2016: 14th {European} {Conference}, {Amsterdam}, {The} {Netherlands}, {October} 11-14, 2016, {Proceedings}, {Part} {II} 14}}, \bibinfo{pages}{869--884} (\bibinfo{publisher}{Springer}, \bibinfo{year}{2016}).

\bibitem{zapletal_vehicle_2016}
\bibinfo{author}{Zapletal, D.} \& \bibinfo{author}{Herout, A.}
\newblock \bibinfo{title}{Vehicle re-identification for automatic video traffic surveillance}.
\newblock In \emph{\bibinfo{booktitle}{Proceedings of the {IEEE} conference on computer vision and pattern recognition workshops}}, \bibinfo{pages}{25--31} (\bibinfo{year}{2016}).

\bibitem{yang_large-scale_2015}
\bibinfo{author}{Yang, L.}, \bibinfo{author}{Luo, P.}, \bibinfo{author}{Loy, C.~C.} \& \bibinfo{author}{Tang, X.}
\newblock \bibinfo{title}{A large-scale car dataset for fine-grained categorization and verification}.
\newblock In \emph{\bibinfo{booktitle}{2015 {IEEE} {Conference} on {Computer} {Vision} and {Pattern} {Recognition} ({CVPR})}}, \bibinfo{pages}{3973--3981}, \doiprefix\url{10.1109/CVPR.2015.7299023} (\bibinfo{year}{2015}).
\newblock \bibinfo{note}{ISSN: 1063-6919}.

\bibitem{nazemi_unsupervised_2018}
\bibinfo{author}{Nazemi, A.}, \bibinfo{author}{Shafiee, M.~J.}, \bibinfo{author}{Azimifar, Z.} \& \bibinfo{author}{Wong, A.}
\newblock \bibinfo{title}{Unsupervised {Feature} {Learning} {Toward} a {Real}-time {Vehicle} {Make} and {Model} {Recognition}}, \doiprefix\url{10.48550/arXiv.1806.03028} (\bibinfo{year}{2018}).
\newblock \bibinfo{note}{ArXiv:1806.03028 [cs]}.

\bibitem{szegedy_going_2015}
\bibinfo{author}{Szegedy, C.} \emph{et~al.}
\newblock \bibinfo{title}{Going deeper with convolutions}.
\newblock In \emph{\bibinfo{booktitle}{2015 {IEEE} {Conference} on {Computer} {Vision} and {Pattern} {Recognition} ({CVPR})}}, \bibinfo{pages}{1--9}, \doiprefix\url{10.1109/CVPR.2015.7298594} (\bibinfo{publisher}{IEEE}, \bibinfo{address}{Boston, MA, USA}, \bibinfo{year}{2015}).

\bibitem{lee_real-time_2019}
\bibinfo{author}{Lee, H.~J.}, \bibinfo{author}{Ullah, I.}, \bibinfo{author}{Wan, W.}, \bibinfo{author}{Gao, Y.} \& \bibinfo{author}{Fang, Z.}
\newblock \bibinfo{journal}{\bibinfo{title}{Real-{Time} {Vehicle} {Make} and {Model} {Recognition} with the {Residual} {SqueezeNet} {Architecture}}}.
\newblock {\emph{\JournalTitle{Sensors}}} \textbf{\bibinfo{volume}{19}}, \bibinfo{pages}{982}, \doiprefix\url{10.3390/s19050982} (\bibinfo{year}{2019}).
\newblock \bibinfo{note}{Number: 5 Publisher: Multidisciplinary Digital Publishing Institute}.

\bibitem{liu_learn_2023}
\bibinfo{author}{Liu, D.}, \bibinfo{author}{Zhao, L.}, \bibinfo{author}{Wang, Y.} \& \bibinfo{author}{Kato, J.}
\newblock \bibinfo{journal}{\bibinfo{title}{Learn from each other to {Classify} better: {Cross}-layer mutual attention learning for fine-grained visual classification}}.
\newblock {\emph{\JournalTitle{Pattern Recognition}}} \textbf{\bibinfo{volume}{140}}, \bibinfo{pages}{109550}, \doiprefix\url{10.1016/j.patcog.2023.109550} (\bibinfo{year}{2023}).

\bibitem{krause_3d_2013}
\bibinfo{author}{Krause, J.}, \bibinfo{author}{Stark, M.}, \bibinfo{author}{Deng, J.} \& \bibinfo{author}{Fei-Fei, L.}
\newblock \bibinfo{title}{{3D} {Object} {Representations} for {Fine}-{Grained} {Categorization}}.
\newblock In \emph{\bibinfo{booktitle}{2013 {IEEE} {International} {Conference} on {Computer} {Vision} {Workshops}}}, \bibinfo{pages}{554--561}, \doiprefix\url{10.1109/ICCVW.2013.77} (\bibinfo{year}{2013}).

\bibitem{yu_night-time_2024}
\bibinfo{author}{Yu, Y.}, \bibinfo{author}{Chen, W.}, \bibinfo{author}{Chen, F.}, \bibinfo{author}{Jia, W.} \& \bibinfo{author}{Lu, Q.}
\newblock \bibinfo{journal}{\bibinfo{title}{Night-time vehicle model recognition based on domain adaptation}}.
\newblock {\emph{\JournalTitle{Multimedia Tools and Applications}}} \textbf{\bibinfo{volume}{83}}, \bibinfo{pages}{9577--9596}, \doiprefix\url{10.1007/s11042-023-15447-1} (\bibinfo{year}{2024}).

\bibitem{tan_coarse--fine_2023}
\bibinfo{author}{Tan, S.~H.}, \bibinfo{author}{Chuah, J.~H.}, \bibinfo{author}{Chow, C.-O.} \& \bibinfo{author}{Kanesan, J.}
\newblock \bibinfo{journal}{\bibinfo{title}{Coarse-to-{Fine} {Context} {Aggregation} {Network} for {Vehicle} {Make} and {Model} {Recognition}}}.
\newblock {\emph{\JournalTitle{IEEE Access}}}  (\bibinfo{year}{2023}).
\newblock \bibinfo{note}{ISBN: 2169-3536 Publisher: IEEE}.

\bibitem{simonyan_very_2014}
\bibinfo{author}{Simonyan, K.} \& \bibinfo{author}{Zisserman, A.}
\newblock \bibinfo{journal}{\bibinfo{title}{Very deep convolutional networks for large-scale image recognition}}.
\newblock {\emph{\JournalTitle{arXiv preprint arXiv:1409.1556}}}  (\bibinfo{year}{2014}).

\bibitem{szegedy_rethinking_2016}
\bibinfo{author}{Szegedy, C.}, \bibinfo{author}{Vanhoucke, V.}, \bibinfo{author}{Ioffe, S.}, \bibinfo{author}{Shlens, J.} \& \bibinfo{author}{Wojna, Z.}
\newblock \bibinfo{title}{Rethinking the inception architecture for computer vision}.
\newblock In \emph{\bibinfo{booktitle}{Proceedings of the {IEEE} conference on computer vision and pattern recognition}}, \bibinfo{pages}{2818--2826} (\bibinfo{year}{2016}).

\bibitem{he_deep_2016}
\bibinfo{author}{He, K.}, \bibinfo{author}{Zhang, X.}, \bibinfo{author}{Ren, S.} \& \bibinfo{author}{Sun, J.}
\newblock \bibinfo{title}{Deep residual learning for image recognition}.
\newblock In \emph{\bibinfo{booktitle}{Proceedings of the {IEEE} conference on computer vision and pattern recognition}}, \bibinfo{pages}{770--778} (\bibinfo{year}{2016}).

\bibitem{huang_densely_2017}
\bibinfo{author}{Huang, G.}, \bibinfo{author}{Liu, Z.}, \bibinfo{author}{Van Der~Maaten, L.} \& \bibinfo{author}{Weinberger, K.~Q.}
\newblock \bibinfo{title}{Densely connected convolutional networks}.
\newblock In \emph{\bibinfo{booktitle}{Proceedings of the {IEEE} conference on computer vision and pattern recognition}}, \bibinfo{pages}{4700--4708} (\bibinfo{year}{2017}).

\bibitem{wu_graph_2022}
\bibinfo{author}{Wu, H.}, \bibinfo{author}{Guo, H.}, \bibinfo{author}{Miao, Q.}, \bibinfo{author}{Huang, M.} \& \bibinfo{author}{Wang, J.}
\newblock \bibinfo{title}{Graph neural networks based multi-granularity feature representation learning for fine-grained visual categorization}.
\newblock In \emph{\bibinfo{booktitle}{International {Conference} on {Multimedia} {Modeling}}}, \bibinfo{pages}{230--242} (\bibinfo{publisher}{Springer}, \bibinfo{year}{2022}).

\bibitem{ali_vehicle_2022}
\bibinfo{author}{Ali, M.}, \bibinfo{author}{Tahir, M.~A.} \& \bibinfo{author}{Durrani, M.~N.}
\newblock \bibinfo{journal}{\bibinfo{title}{Vehicle images dataset for make and model recognition}}.
\newblock {\emph{\JournalTitle{Data in brief}}} \textbf{\bibinfo{volume}{42}} (\bibinfo{year}{2022}).
\newblock \bibinfo{note}{ISBN: 2352-3409 Publisher: Elsevier}.

\bibitem{he_transfg_2022}
\bibinfo{author}{He, J.} \emph{et~al.}
\newblock \bibinfo{title}{Transfg: {A} transformer architecture for fine-grained recognition}.
\newblock In \emph{\bibinfo{booktitle}{Proceedings of the {AAAI} conference on artificial intelligence}}, vol.~\bibinfo{volume}{36}, \bibinfo{pages}{852--860} (\bibinfo{year}{2022}).
\newblock \bibinfo{note}{Issue: 1}.

\bibitem{pedro}
\bibinfo{author}{Felzenszwalb, P.~F.}, \bibinfo{author}{Girshick, R.~B.}, \bibinfo{author}{McAllester, D.} \& \bibinfo{author}{Ramanan, D.}
\newblock \bibinfo{journal}{\bibinfo{title}{Object detection with discriminatively trained part-based models}}.
\newblock {\emph{\JournalTitle{IEEE Transactions on Pattern Analysis and Machine Intelligence}}} \textbf{\bibinfo{volume}{32}}, \bibinfo{pages}{1627--1645} (\bibinfo{year}{2010}).

\bibitem{kubilius2019brainlike}
\bibinfo{author}{Kubilius, J.} \emph{et~al.}
\newblock \bibinfo{journal}{\bibinfo{title}{Brain-like object recognition with high-performing shallow recurrent anns}}.
\newblock {\emph{\JournalTitle{Advances in Neural Information Processing Systems}}} \textbf{\bibinfo{volume}{32}} (\bibinfo{year}{2019}).

\bibitem{dicarlo2012brain}
\bibinfo{author}{DiCarlo, J.~J.}, \bibinfo{author}{Zoccolan, D.} \& \bibinfo{author}{Rust, N.~C.}
\newblock \bibinfo{journal}{\bibinfo{title}{How does the brain solve visual object recognition?}}
\newblock {\emph{\JournalTitle{Neuron}}} \textbf{\bibinfo{volume}{73}}, \bibinfo{pages}{415--434} (\bibinfo{year}{2012}).

\bibitem{ren2015faster}
\bibinfo{author}{Ren, S.}, \bibinfo{author}{He, K.}, \bibinfo{author}{Girshick, R.} \& \bibinfo{author}{Sun, J.}
\newblock \bibinfo{title}{Faster r-cnn: Towards real-time object detection with region proposal networks}.
\newblock In \emph{\bibinfo{booktitle}{Advances in Neural Information Processing Systems}}, \bibinfo{pages}{91--99} (\bibinfo{year}{2015}).

\bibitem{liu2016ssd}
\bibinfo{author}{Liu, W.} \emph{et~al.}
\newblock \bibinfo{title}{Ssd: Single shot multibox detector}.
\newblock In \emph{\bibinfo{booktitle}{European Conference on Computer Vision}}, \bibinfo{pages}{21--37} (\bibinfo{publisher}{Springer}, \bibinfo{year}{2016}).

\bibitem{girshick2015fast}
\bibinfo{author}{Girshick, R.}
\newblock \bibinfo{title}{Fast r-cnn}.
\newblock In \emph{\bibinfo{booktitle}{Proceedings of the IEEE International Conference on Computer Vision}}, \bibinfo{pages}{1440--1448} (\bibinfo{year}{2015}).

\bibitem{uijlings2013selective}
\bibinfo{author}{Uijlings, J.~R.}, \bibinfo{author}{van~de Sande, K.~E.}, \bibinfo{author}{Gevers, T.} \& \bibinfo{author}{Smeulders, A.~W.}
\newblock \bibinfo{journal}{\bibinfo{title}{Selective search for object recognition}}.
\newblock {\emph{\JournalTitle{International Journal of Computer Vision}}} \textbf{\bibinfo{volume}{104}}, \bibinfo{pages}{154--171} (\bibinfo{year}{2013}).

\bibitem{buschman2009serial}
\bibinfo{author}{Buschman, T.~J.} \& \bibinfo{author}{Miller, E.~K.}
\newblock \bibinfo{journal}{\bibinfo{title}{Serial, covert shifts of attention during visual search are reflected by the frontal eye fields and correlated with population oscillations}}.
\newblock {\emph{\JournalTitle{Neuron}}} \textbf{\bibinfo{volume}{63}}, \bibinfo{pages}{386--396} (\bibinfo{year}{2009}).

\bibitem{redmon2016you}
\bibinfo{author}{Redmon, J.}, \bibinfo{author}{Divvala, S.}, \bibinfo{author}{Girshick, R.} \& \bibinfo{author}{Farhadi, A.}
\newblock \bibinfo{title}{You only look once: Unified, real-time object detection}.
\newblock In \emph{\bibinfo{booktitle}{Proceedings of the IEEE Conference on Computer Vision and Pattern Recognition}}, \bibinfo{pages}{779--788} (\bibinfo{year}{2016}).

\bibitem{redmon2018yolov3}
\bibinfo{author}{Redmon, J.} \& \bibinfo{author}{Farhadi, A.}
\newblock \bibinfo{journal}{\bibinfo{title}{Yolov3: An incremental improvement}}.
\newblock {\emph{\JournalTitle{arXiv preprint arXiv:1804.02767}}}  (\bibinfo{year}{2018}).

\bibitem{fahle1994human}
\bibinfo{author}{Fahle, M.}
\newblock \bibinfo{journal}{\bibinfo{title}{Human pattern recognition: parallel processing and perceptual learning}}.
\newblock {\emph{\JournalTitle{Perception}}} \textbf{\bibinfo{volume}{23}}, \bibinfo{pages}{411--427} (\bibinfo{year}{1994}).

\bibitem{treisman1980feature}
\bibinfo{author}{Treisman, A.~M.} \& \bibinfo{author}{Gelade, G.}
\newblock \bibinfo{journal}{\bibinfo{title}{A feature-integration theory of attention}}.
\newblock {\emph{\JournalTitle{Cognitive psychology}}} \textbf{\bibinfo{volume}{12}}, \bibinfo{pages}{97--136} (\bibinfo{year}{1980}).

\bibitem{wertheimer1938gestalt}
\bibinfo{author}{Wertheimer, M.}
\newblock \bibinfo{title}{Laws of organization in perceptual forms}.
\newblock In \bibinfo{editor}{Ellis, W.~D.} (ed.) \emph{\bibinfo{booktitle}{A Source Book of Gestalt Psychology}}, \bibinfo{pages}{71--88} (\bibinfo{publisher}{Routledge \& Kegan Paul}, \bibinfo{year}{1938}).
\newblock \bibinfo{note}{Originally published in 1923 as ``Untersuchungen zur Lehre von der Gestalt II''}.

\bibitem{neill1987selective}
\bibinfo{author}{Neill, W.~T.} \& \bibinfo{author}{Westberry, R.~L.}
\newblock \bibinfo{journal}{\bibinfo{title}{Selective attention and the suppression of cognitive noise.}}
\newblock {\emph{\JournalTitle{Journal of Experimental Psychology: Learning, Memory, and Cognition}}} \textbf{\bibinfo{volume}{13}}, \bibinfo{pages}{327} (\bibinfo{year}{1987}).

\bibitem{yolo11_ultralytics}
\bibinfo{author}{Jocher, G.} \& \bibinfo{author}{Qiu, J.}
\newblock \bibinfo{title}{Ultralytics yolo11} (\bibinfo{year}{2024}).

\bibitem{cheng2024yolo}
\bibinfo{author}{Cheng, T.} \emph{et~al.}
\newblock \bibinfo{title}{Yolo-world: Real-time open-vocabulary object detection}.
\newblock In \emph{\bibinfo{booktitle}{Proceedings of the IEEE/CVF Conference on Computer Vision and Pattern Recognition}}, \bibinfo{pages}{16901--16911} (\bibinfo{year}{2024}).

\bibitem{COCO}
\bibinfo{author}{Lin, T.-Y.} \emph{et~al.}
\newblock \bibinfo{journal}{\bibinfo{title}{Microsoft {COCO:} common objects in context}}.
\newblock {\emph{\JournalTitle{CoRR}}} \textbf{\bibinfo{volume}{abs/1405.0312}} (\bibinfo{year}{2014}).
\newblock \eprint{1405.0312}.

\bibitem{lvis}
\bibinfo{author}{Gupta, A.}, \bibinfo{author}{Dollar, P.}, \bibinfo{author}{Girshick, R.} \emph{et~al.}
\newblock \bibinfo{title}{{LVIS}: A dataset for large vocabulary instance segmentation}.
\newblock In \emph{\bibinfo{booktitle}{Proceedings of the IEEE Conference on Computer Vision and Pattern Recognition}} (\bibinfo{year}{2019}).

\bibitem{ATSS}
\bibinfo{author}{Zhang, S.}, \bibinfo{author}{Chi, C.}, \bibinfo{author}{Yao, Y.}, \bibinfo{author}{Lei, Z.} \& \bibinfo{author}{Li, S.~Z.}
\newblock \bibinfo{title}{Bridging the gap between anchor-based and anchor-free detection via adaptive training sample selection}.
\newblock In \emph{\bibinfo{booktitle}{Proceedings of the IEEE/CVF Conference on Computer Vision and Pattern Recognition}}, \bibinfo{pages}{9756--9765} (\bibinfo{year}{2020}).

\bibitem{DINO}
\bibinfo{author}{Zhang, H.} \emph{et~al.}
\newblock \bibinfo{journal}{\bibinfo{title}{Dino: Detr with improved denoising anchor boxes for end-to-end object detection}}.
\newblock {\emph{\JournalTitle{arXiv preprint arXiv:2203.03605}}}  (\bibinfo{year}{2022}).

\bibitem{CLIP}
\bibinfo{author}{Radford, A.} \emph{et~al.}
\newblock \bibinfo{journal}{\bibinfo{title}{Learning transferable visual models from natural language supervision}}.
\newblock {\emph{\JournalTitle{arXiv preprint arXiv:2103.00020}}}  (\bibinfo{year}{2021}).

\bibitem{GLIP}
\bibinfo{author}{Li, J.}, \bibinfo{author}{Li, D.}, \bibinfo{author}{Xiong, C.} \& \bibinfo{author}{Hoi, S.}
\newblock \bibinfo{journal}{\bibinfo{title}{Glip: Grounded language-image pre-training}}.
\newblock {\emph{\JournalTitle{arXiv preprint arXiv:2112.03857}}}  (\bibinfo{year}{2022}).

\bibitem{vstancel2019introduction}
\bibinfo{author}{{\v{S}}tancel, M.} \& \bibinfo{author}{Huli{\v{c}}, M.}
\newblock \bibinfo{title}{An introduction to image classification and object detection using yolo detector}.
\newblock In \emph{\bibinfo{booktitle}{CEUR workshop proceedings}}, vol. \bibinfo{volume}{2403}, \bibinfo{pages}{1--8} (\bibinfo{year}{2019}).

\bibitem{jiang2022review}
\bibinfo{author}{Jiang, P.}, \bibinfo{author}{Ergu, D.}, \bibinfo{author}{Liu, F.}, \bibinfo{author}{Cai, Y.} \& \bibinfo{author}{Ma, B.}
\newblock \bibinfo{journal}{\bibinfo{title}{A review of yolo algorithm developments}}.
\newblock {\emph{\JournalTitle{Procedia computer science}}} \textbf{\bibinfo{volume}{199}}, \bibinfo{pages}{1066--1073} (\bibinfo{year}{2022}).

\bibitem{friston2005theory}
\bibinfo{author}{Friston, K.}
\newblock \bibinfo{journal}{\bibinfo{title}{A theory of cortical responses}}.
\newblock {\emph{\JournalTitle{Philosophical Transactions of the Royal Society B: Biological Sciences}}} \textbf{\bibinfo{volume}{360}}, \bibinfo{pages}{815--836} (\bibinfo{year}{2005}).

\bibitem{richardson2006memory}
\bibinfo{author}{Richardson-Klavehn, A.} \& \bibinfo{author}{Bjork, R.~A.}
\newblock \bibinfo{journal}{\bibinfo{title}{Memory and the self in cognitive neuroscience}}.
\newblock {\emph{\JournalTitle{Cognition}}} \textbf{\bibinfo{volume}{99}}, \bibinfo{pages}{123--132} (\bibinfo{year}{2006}).

\bibitem{tyler2013objects}
\bibinfo{author}{Tyler, L.~K.} \emph{et~al.}
\newblock \bibinfo{journal}{\bibinfo{title}{Objects and categories: feature statistics and object processing in the ventral stream}}.
\newblock {\emph{\JournalTitle{Journal of cognitive neuroscience}}} \textbf{\bibinfo{volume}{25}}, \bibinfo{pages}{1723--1735} (\bibinfo{year}{2013}).

\bibitem{brojde2012bilingual}
\bibinfo{author}{Brojde, C.~L.}, \bibinfo{author}{Ahmed, S.} \& \bibinfo{author}{Colunga, E.}
\newblock \bibinfo{journal}{\bibinfo{title}{Bilingual and monolingual children attend to different cues when learning new words}}.
\newblock {\emph{\JournalTitle{Frontiers in Psychology}}} \textbf{\bibinfo{volume}{3}}, \bibinfo{pages}{155} (\bibinfo{year}{2012}).

\bibitem{anpr_survey}
\bibinfo{author}{Du, S.}, \bibinfo{author}{Ibrahim, M.}, \bibinfo{author}{Shehata, M.} \& \bibinfo{author}{Badawy, W.}
\newblock \bibinfo{journal}{\bibinfo{title}{Automatic number plate recognition (anpr): A review}}.
\newblock {\emph{\JournalTitle{IEEE Transactions on Intelligent Transportation Systems}}} \textbf{\bibinfo{volume}{16}}, \bibinfo{pages}{2233--2249} (\bibinfo{year}{2013}).

\bibitem{yolov8}
\bibinfo{author}{Ultralytics}.
\newblock \bibinfo{title}{Yolov8: Ultralytics official implementation}.
\newblock \bibinfo{howpublished}{\url{https://github.com/ultralytics/ultralytics}} (\bibinfo{year}{2023}).
\newblock \bibinfo{note}{Available at \url{https://docs.ultralytics.com/}}.

\end{thebibliography}

\end{document}